\documentclass{article}

\usepackage{microtype}
\usepackage{graphicx}
\usepackage{subfigure}
\usepackage{booktabs} 

\usepackage{amsmath,amsfonts,amssymb,amsthm,dsfont,stmaryrd}
\usepackage[utf8]{inputenc}
\usepackage{hyperref}

\usepackage{graphicx}

\usepackage{amssymb}

\usepackage{color}
\usepackage{multirow}

\usepackage{color}
\def\R{\mathbb{R}}

\def\our{FlowSVDD}

\usepackage{hyperref}

\usepackage[accepted]{icml2020}

\icmltitlerunning{Flow-based SVDD for anomaly detection}

\begin{document}

\twocolumn[
\icmltitle{Flow-based SVDD for anomaly detection}




\begin{icmlauthorlist}
\icmlauthor{Marcin Sendera}{jagiellonian}
\icmlauthor{Marek Śmieja}{jagiellonian}
\icmlauthor{Łukasz Maziarka}{jagiellonian}
\icmlauthor{Łukasz Struski}{jagiellonian}
\icmlauthor{Przemysław Spurek}{jagiellonian}
\icmlauthor{Jacek Tabor}{jagiellonian}

\end{icmlauthorlist}

\icmlaffiliation{jagiellonian}{Faculty of Mathematics and Computer
Science, Jagiellonian University, Kraków, Poland}

\icmlcorrespondingauthor{Marcin Sendera}{marcin.sendera@gmail.com}

\icmlkeywords{Machine Learning, ICML}

\vskip 0.3in
]



\printAffiliationsAndNotice{\icmlEqualContribution} 

\begin{abstract}
We propose \our{} -- a flow-based one-class classifier for anomaly/outliers detection that realizes a well-known SVDD principle using deep learning tools. Contrary to other approaches to deep SVDD, the proposed model is instantiated using flow-based models, which naturally prevents from collapsing of bounding hypersphere into a single point. Experiments show that \our{} achieves comparable results to the current state-of-the-art methods and significantly outperforms related deep SVDD methods on benchmark datasets.

\end{abstract}

\section{Introduction}

Anomaly (novelty/outlier) detection refers to the identification of novel or abnormal patterns embedded in a large amount of typical (normal) data \cite{miljkovic2010review}.
Anomaly detection algorithms find application in fraud detection systems, discovering failures in the industrial domain, detection of adversarial examples, etc.. 

In contrast to typical binary classification problems, where every class follows some probability distribution, an anomaly is a pattern that does not conform to the expected behavior. In consequence, a completely novel type of anomalies can occur at test time, which is not similar to any known anomalies. Moreover, in most cases, we do not have access to any anomalies at training time. In consequence, novelty detection is usually solved using unsupervised approaches, such as one-class classifiers, which focus on describing the behavior of available data (inliers). Any observation, which deviates from this behavior, is labeled as an outlier. 

Our research is motivated by the idea of Support Vector Data Description (SVDD) \cite{tax2004support}, which obtains a spherically shaped boundary around a dataset by usage of soft margin and penalization of data points from outside the bounding region. We propose \our{} -- a one-class classifier based on flow-based models \cite{dinh2014nice}, which finds a hypersphere with a minimal volume that encloses data. Since flow-based models are commonly used in the context of generative models, we redefine their cost function to minimize the volume of the bounding hypersphere instead of maximizing the log-likelihood function. On one hand, flow-based models allow us to calculate a Jacobian of a neural network at every point. In consequence, minimizing the volume of the hypersphere in the feature space leads to the minimization of the volume of the corresponding bounding region in the input space. On the other hand, since flow-based models give an explicit formula for the inverse mapping, we automatically get a parametric form for the corresponding bounding region in the input space. In contrast to deep SVDD models, our approach eliminates the problem of hypersphere collapse, which makes it easy to use. 

Extensive experiments performed on typical benchmark datasets show that our method significantly outperforms the deep SVDD model while being comparative to state-of-the-art models for anomaly detection.

Our contribution is summarized as follows:
\begin{enumerate}
    \item We propose an adaptation of the SVDD method to deep neural networks with the use of flow models.
    \item We show that the realization of the SVDD loss function on flow-based models prevents from hypersphere collapse.
    \item We experimentally compare \our{} with Deep SVDD and current state-of-the-art methods.
\end{enumerate}

\section{Proposed model}

\paragraph{Preliminaries: SVDD.} 
Our approach is motivated by a classical Support Vector Data Description (SVDD) \cite{tax2004support}, which tries to find a minimal hypersphere to enclose the data. To allow the possibility of outliers in the training set, SVDD uses a soft margin and penalizes data points that lie outside the bounding hypersphere. If $f$ maps input data to the output kernel space, then SVDD loss equals:
\begin{equation} \label{eq:svdd}
F(R,c;f)=R^2+\frac{1}{\nu n} \sum_{i} \max(0,\|f(x_i)-c\|^2-R^2)
\end{equation}
where $c \in \R^D, R \in \R$ is the center and the radius of the hypersphere, respectively, and $\nu$ is the trade-off between the volume and boundary violations of the hypersphere, i.e. fraction of outliers.

The realization of SVDD using deep neural networks was presented in \cite{ruff2018deep} (it was termed DSVDD). However, direct minimization of the SVDD loss may lead to a trivial solution, i.e. the hypersphere collapses to a single point $c$. To avoid this negative behavior, it has been recommended that the center $c$ must be something other than the all-zero-weights solution, and the network should use only unbounded activations and omit bias terms. While the two first conditions can be accepted, omitting bias terms in a network may lead to a sub-optimal feature representation due to the role of bias in shifting activation values. 

To eliminate the above restrictions a recent work \cite{chong2020simple} proposes two regularizers, which prevent hypersphere collapse, and uses an adaptive weighting scheme to control the amount of penalization between the SVDD loss and the respective regularizer.

\paragraph{Flow-based SVDD.} 
As an alternative to DSVDD, we realize the SVDD objective using flow models. Let us recall that a neural network $f:\R^D \to \R^D$ is a flow model if the inverse mapping $f^{-1}$ is given explicitly and the Jacobian determinant $w(x) = \det d f(x)$ can be easily calculated. In our approach, we use a special class of flow models, in which Jacobian determinant is constant at every point, i.e. $w(x) = w$, such as NICE \cite{dinh2014nice}. In this case, we get a natural correspondence between the volume of the bounding hypersphere in the output space and the volume of a bounding region in the input space, see below. 

Let us first consider the simplest situation when $w = 1$. In such a scenario, the volume of any shape in the input space equals the volume of its image in the output feature space. In consequence, a direct minimization of the SVDD objective does not lead to the hypersphere collapse.

In a more general scenario, when $w \neq 1$, we need to include $w$ in the SVDD objective. Observe that the  Jacobian determinant of the mapping $f/w^{1/D}$ equals 1. Thus to get the equality of the volume in the input and output space, we redefine the SVDD loss \eqref{eq:svdd} as follows:

\begin{equation} \label{eq:our}
F(R,c;f)=R^2+\frac{1}{\nu n}\sum_{i=1}^n \max(0,\|f(x_i)/w^{1/D}-c\|^2-R^2).
\end{equation}

In a test phase, a given example $x$ is deemed as an outlier if: 
$$
\|f(x)/w^{1/D}-c\| > R,
$$
which is equivalent to 
$$
\|f(x)-w^{1/D}c\| >R w^{1/D}.
$$
In other words, inliers lie inside the ball $B(w^{1/D}c; Rw^{1/D})$.

\section{Experiments}

In this section, we experimentally examine \our{} and compare it with several state-of-the-art approaches. \our{} is implemented using the architecture of the NICE flow model  (4 coupling layers -- each consisted of 4 layers and 256 hidden dimensions) with constant Jacobian determinant and $\nu = 0.05$.

\begin{figure}[H] 
    \centering
    \includegraphics[width=0.23\textwidth]{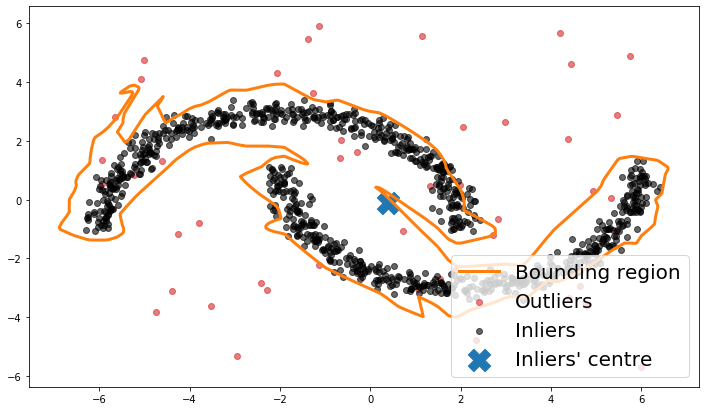}
    \includegraphics[width=0.23\textwidth]{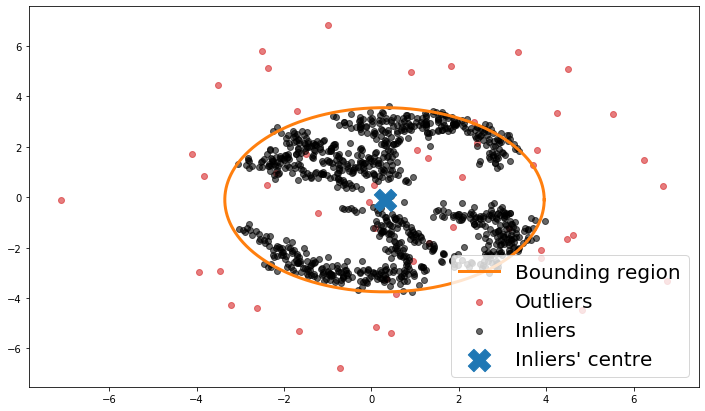}
    
    \includegraphics[width=0.23\textwidth]{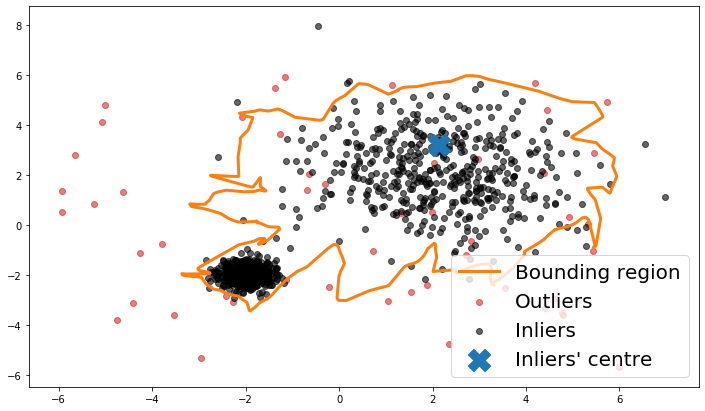}
    \includegraphics[width=0.23\textwidth]{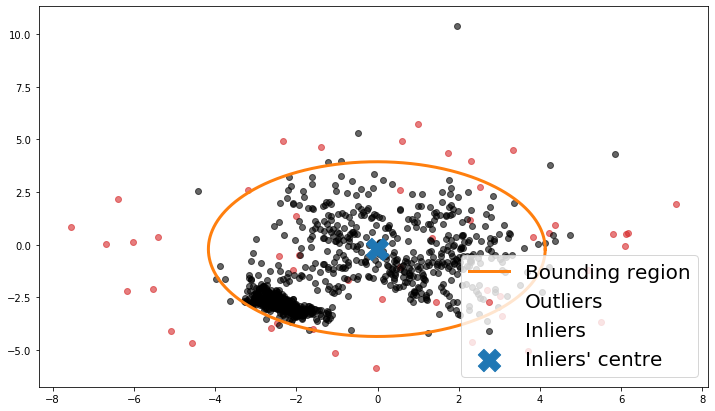}
    \caption{Enclosing hyperspheres in the latent space of \our{} (right) and the corresponding bounding regions in the input space (left).}
    \label{fig:plots_2d}
\end{figure}

\paragraph{Illustrative example.} 
To get the intuition behind \our{}, we first consider 2-dimensional examples, which are easy to visualize. The results presented in Figure \ref{fig:plots_2d} show the resulting hyperspheres in the latent space and the corresponding bounding regions in the input space. At first glance, we can observe that the bounding region in the original space is close to the structure of inliers. In the latent space, \our{} finds the center point $c$ and radius $R$ to enclose $(1-\nu)$ percentage of data inside the ball $B(c; R)$. Observe that, unlike the density-based flow models, \our{} does not transform data into Gaussian distribution in a latent space.

\paragraph{Benchmark  data  for  anomaly  detection.}
To provide quantitative assessment, we take into account Thyroid\footnote{\url{http://odds.cs.stonybrook.edu/thyroid-disease-dataset/}} and KDDCUP\footnote{\url{http://kdd.ics.uci.edu/databases/kddcup99/kddcup.testdata.unlabeled_10_percent.gz}} datasets, which are typically used for anomaly detection. We use the standard training and test splits and follow exactly the same evaluation protocol as in \cite{wang2019multivariate}. In particular, we use the F1 score and the Area Under Receiver Operating Characteristic curve  (AUC).

Our model is compared with the following algorithms: (1) One-class SVM (OC-SVM) \cite{scholkopf2001estimating}, (2) Deep structured energy-based models (DSEBM) \cite{zhai2016deep}, (3) Deep autoencoding Gaussian mixture model (DAGMM) \cite{zong2018deep}, (4) variants of MQT -- multivariate quantile map (NLL, TQM$_1$, TQM$_2$, TQM$_\infty$) \cite{wang2019multivariate} and (5) Deep Support Vector Data Description (DSVDD) \cite{ruff2018deep} - another implementation of SVDD cost function in deep neural networks.

\begin{table*}[h!]
\caption{Performance on two anomaly detection datasets.} \label{tab:res1}
\centering
\begin{tabular}{lccccccccc }
\toprule
\multicolumn{10}{c}{Thyroid}\\
\midrule
 &  OC-SVM  &  DSEBM  &  DAGMM  &    NLL  &  TQM$_{1}$ &  TQM$_2$  &  TQM$_{\infty}$  &  DSVDD &  \our{}  \\
\midrule
 F1 & .3887 & .0403 & .4782 &  .7312 & .5269 &  .5806 & .7527 & - & .7097 \\
 AUC &       -  &      -  &      -  &       -  &     - &    -  &    -  & 0.749 & .9797 \\
\midrule
 \multicolumn{10}{c}{KDDCUP}\\
\midrule
 &  OC-SVM  &  DSEBM  &  DAGMM  &     NLL  &  TQM$_{1}$ &  TQM$_2$  &  TQM$_{\infty}$  &  DSVDD &  \our{}  \\
\midrule
 F1 &    .7954 &   .7423 &   .9369 &   .9622 & .9621 & .9622 & .9622 & - & .9030 \\
 AUC &        - &      -  &      -  &        -  &    -  &    -  &    -  &  -  & .9384 \\
\bottomrule
\end{tabular}
\end{table*}

The results presented in Table \ref{tab:res1} show that \our{} model performs better than most methods on the Thyroid dataset and is significantly better than DSVDD in terms of the AUC metric. In the case of KDDCUP, \our{} achieves a score in between the classical methods and current state-of-the-art.

\paragraph{Image datasets.} 
To provide further experimental verification, we use two image datasets: MNIST and Fashion-MNIST. In contrast to the previous comparison, these two datasets are usually used for multiclass classification and thus need to be adapted to the problem of anomaly detection. For this purpose, each of the ten classes is deemed as the nominal class while the rest of the nine classes are deemed as the anomaly class, which results in 10 scenarios for each dataset.

We additionally compare \our{} with the following models: (1) Geometric transformation (GT) \cite{golan2018deep}, Variational autoencoder (VAE) \cite{kingma2013auto}, Denoising autoencoder (DAE) \cite{vincent2008extracting}, Generative probabilistic novelty detection (GPND) \cite{pidhorskyi2018generative}, Latent space autoregression (LSA) \cite{abati2019latent}. In contrast to previous experiment, we only use TQM$_2$ and NLL as the only implementations of MTQ, because they output the highest value of AUC \cite{wang2019multivariate}.

\begin{figure*}[h!] 
    \centering
        \includegraphics[width=0.48\textwidth]{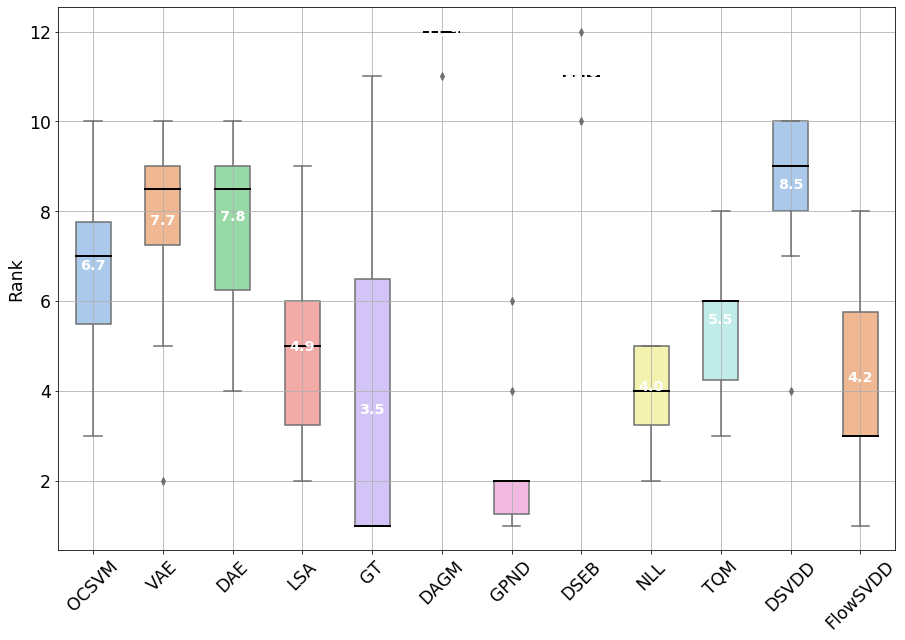}
        \label{fig:ranks_mnist}
        \includegraphics[width=0.48\textwidth]{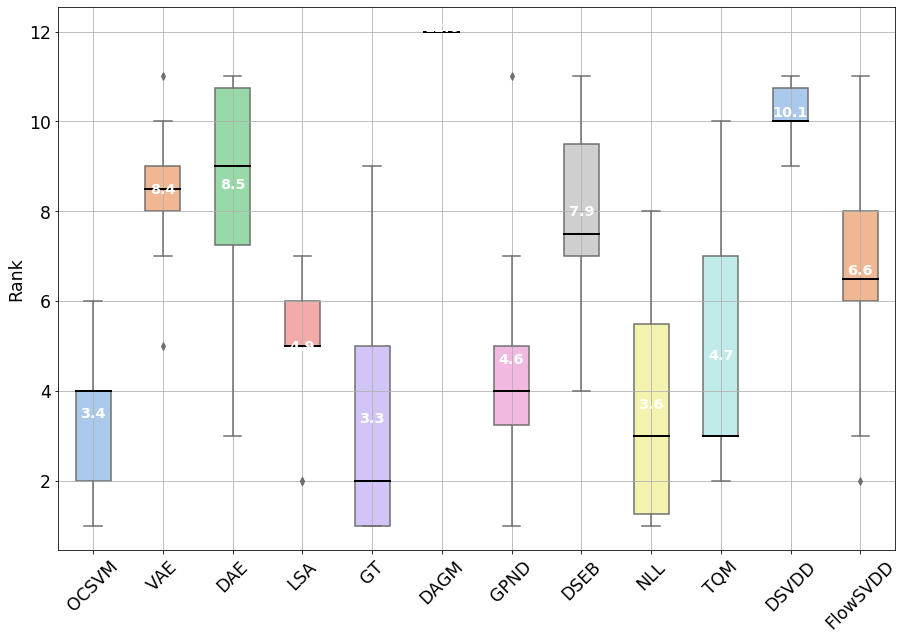}
        \label{fig:ranks_fmnist}
    \caption{Box plots for rankings calculated on MNIST (left) and Fashion-MNIST (right). The median ranking is marked by a line, while the average ranking is marked with a number.}
    \label{fig:ranks}
\end{figure*}

To present the results, we compute the ranking on each of 10 scenarios and summarize it using a box plot, see Figure \ref{fig:ranks}.  The results show that \our{} significantly outperforms DSVDD in both datasets. In the case of the MNIST dataset, we observe that \our{} is almost as good as the current state-of-the-art methods, like GT and NLL.

\begin{figure}[H] 
    \centering
    \includegraphics[width=0.49\textwidth]{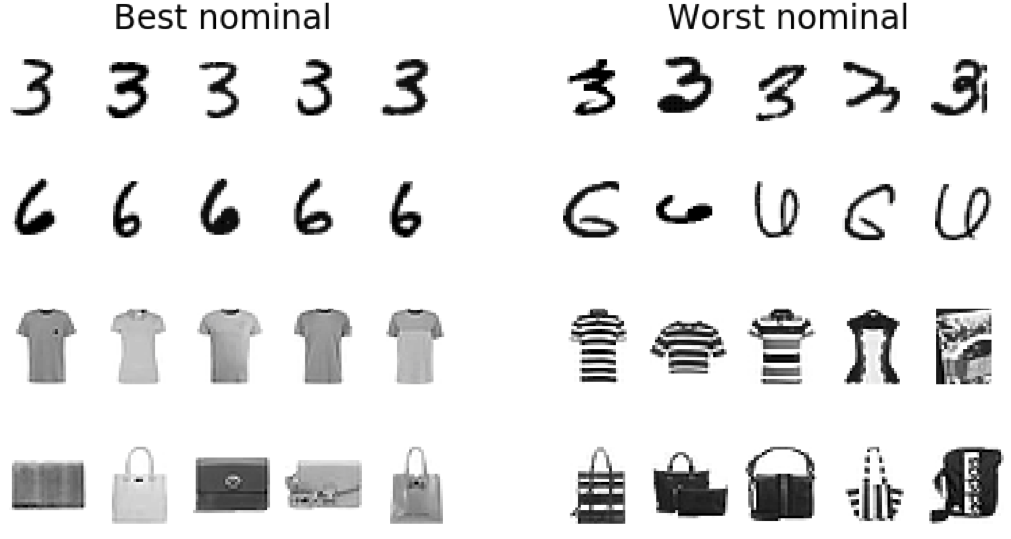}
    \caption{Best  nominal  (left)  and  worst  nominal (right) examples determined by \our{} for MNIST (top) and Fashion-MNIST (bottom).}
    \label{fig:samples}
\end{figure}

Finally, we analyze, which samples are localized close to or furthest from the center of bounding hypersphere. 
Results in Figure \ref{fig:samples} shows that \our{} maps regular images in the hypersphere center. Contrary, examples localized far from the center, could be hard to identify.
It means that \our{} gives results consistent with our intuition.

\section{Conclusion}

The paper introduced \our{}, which realizes the SVDD paradigm in the case of neural networks. Making use of flow-based models and an appropriate SVDD-like cost function, we find a minimal bounding region for a majority of data.  Unlike other deep SVDD realizations, \our{} does not change the determinant of a Jacobian matrix, which means that the resulting hypersphere cannot collapse in a latent space. The experimental results demonstrate that \our{} presents a very good performance in the case of both artificial and real-world one-class settings.

\section*{Acknowledgements}
The work of M. Śmieja was supported by the National Centre of Science (Poland) Grant No. 2018/31/B/ST6/00993.
The work of P. Spurek was supported by the National Centre of Science (Poland) Grant No. 2019/33/B/ST6/00894. The work of J. Tabor was supported by the National Centre of Science (Poland) Grant No. 2017/25/B/ST6/01271.

\bibliography{icml-workshop-paper}
\bibliographystyle{icml2020}

\end{document}